\documentclass[a4paper]{article}

\usepackage{INTERSPEECH2021}
\usepackage{url}
\usepackage{booktabs}
\usepackage{multirow}
\usepackage{adjustbox}
\usepackage{xcolor}
\usepackage{caption}
\usepackage{siunitx}
\usepackage{textcomp}
\usepackage{subcaption}

\title{Lexical Tone is Hard to Quantize: Probing Discrete Speech Units in Mandarin and Yorùbá}
\name{Opeyemi Osakuade, Simon King}
\address{The Centre for Speech Technology Research, University of Edinburgh, UK}
\email{
O.M.Osakuade@sms.ed.ac.uk, Simon.King@ed.ac.uk}

\usepackage{siunitx}
\usepackage{xspace}
\def\BibTeX{{\rm B\kern-.05em{\sc i\kern-.025em b}\kern-.08em
    T\kern-.1667em\lower.7ex\hbox{E}\kern-.125emX}}
    
\def\F0{$F_0$\xspace}

\begin{document}

\maketitle
\begin{abstract} 
Discrete speech units (DSUs) are derived by quantising representations from models trained using self-supervised learning (SSL). They are a popular representation for a wide variety of spoken language tasks, including those where prosody matters. DSUs are especially convenient for tasks where text and speech are jointly modelled, such as text-to-speech and multimodal dialogue systems. But we have found that DSUs encode suprasegmental information less reliably than segmental structure, which we demonstrate in this work using lexical tone, though this limitation likely extends to other suprasegmental features such as prosody.


Our investigations using the tone languages Mandarin and Yorùbá show that the SSL latent representations themselves do encode tone, yet DSUs obtained using quantisation tend to prioritise phonetic structure, which makes lexical tone less reliably encoded. This remains true for a variety of quantisation methods, not only the most common, K-means.


We conclude that current DSU quantisation strategies have limitations for suprasegmental features, which suggests a need for new, tone-aware (or prosody-aware) techniques in speech representation learning. We point towards a potential form of solution by performing K-means clustering once to encode phonetic information, then again on the residual representation, which better encodes lexical tone.

\end{abstract}


\section{Introduction}

Self-supervised learning (SSL) has become a key component of many speech processing systems, providing rich latent representations that encode phonetic, lexical, and prosodic information \cite{pouw2024perception,choi24b_interspeech,wallbridge25_interspeech}. To use these continuous representations in downstream tasks, it is often necessary to discretise them into Discrete Speech Units (DSUs). DSUs are a symbolic representation of speech that is compatible with language models and facilitates mixing speech and text representations within a single sequence \cite{shon24_interspeech, chou2023toward}. For language modelling, it is preferable to keep the vocabulary size (i.e., the number of unique DSUs) as small as possible \cite{labrak2025empirical}. DSUs are known to normalise away some unwanted speaker and channel variability while preserving phonetic content \cite{baevski2020wav2vec, hsu2021hubert}. They are widely used
in TTS \cite{polyak21_interspeech, wang2023neuralcodec}, speech-to-speech translation \cite{ popuri22_interspeech, li2023textless}, and spoken language modelling \cite{nguyen2023generative,cui2025recent}


\subsection{Discretisation degrades the representation of tone}

Unfortunately, discretisation can remove information that would be useful to the downstream task. ToneUnit \cite{tao2024toneunit} showed that, even though lexical tone is present in SSL latents, it is poorly preserved after quantisation with K-means. This impacts modelling for tone languages. 

In this work, we focus on two typologically distinct tone languages: Mandarin and Yorùbá. Mandarin has contour tones, while Yorùbá primarily uses level tones with both lexical and grammatical functions.
In both languages, our findings are consistent with prior work \cite{tao2024toneunit,shen2024encoding,osakuade2024discrete}
: SSL latents encode phonetic \textit{and} tonal information, but DSUs (obtained by quantising those latents) are dominated by the phonetic information, representing tone less well than the latents. We begin by contributing further evidence that SSL latents encode both phonetic and tonal information and aim to answer two linked questions: (i) why does quantisation degrade tone information? (ii) could an improved quantisation method mitigate this degradation? 


Whilst not part of this work, we believe that our findings for tone information will naturally also extend to supra-segmental information such as prosody.

\begin{figure*}[t!]
    \centering
    \includegraphics[width=\textwidth]{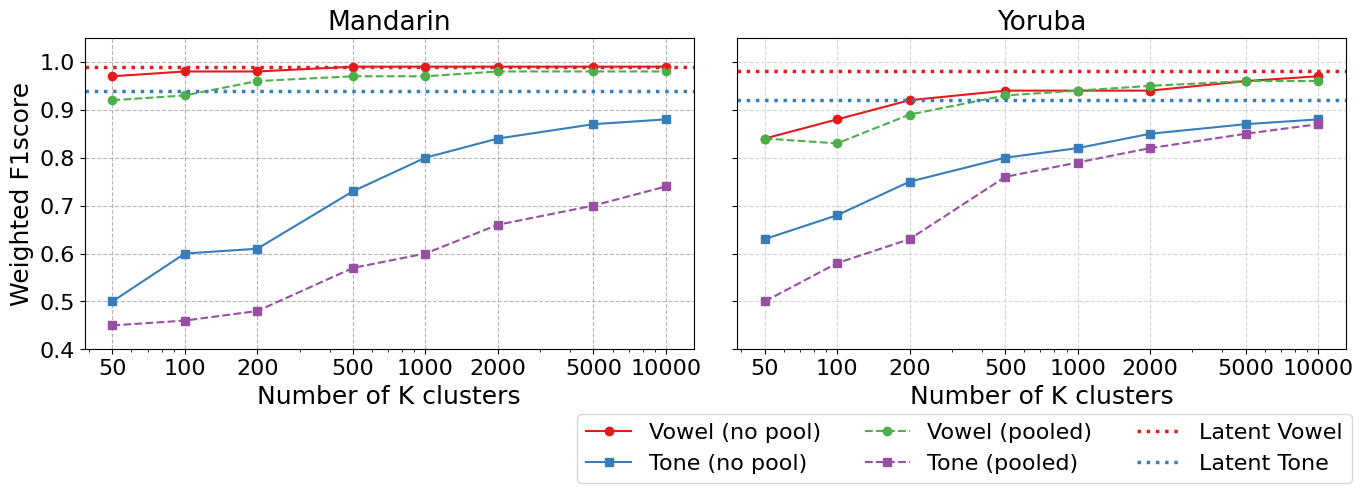} 
    \caption{Weighted F1 scores for Mandarin and Yorùbá phone and tone classification using K-Means codebooks of varying sizes. 
    Solid lines represent no pooling (phone segment-level), while dashed lines represent pooled (phone segment-level) features. 
    Dotted lines denote the corresponding HuBERT latent baselines. The x-axis is logarithmic to highlight performance gains across orders of magnitude in codebook size.}
    \label{fig:kmeans}
\end{figure*}

\subsection{Can better quantisation help?}
\label{sec:motivation}
Lexical tone is a suprasegmental feature realised over the time course of a phone or syllable by a relative change in acoustic correlates, including F0; this is in contrast to the more `instantaneous' and `absolute' nature of  contrasts between vowels or consonants \cite{yip2002tone,fromkin1978tone,burnham2008perception}. Perception by listeners of the acoustic correlates of tone (F0 height and  contour) requires temporal context. Training models using SSL generally involves a reconstruction task (e.g., masked prediction), given the surrounding context. This is presumably why representations obtained from such models (SSL latents) encode tone well \cite{pouw2024perception}. The degradation of tone information occurs when they are discretised. A popular choice is K-means clustering, which treats frames independently. It is self-evident that increasing $K$ will reduce the loss of information, but this has quickly diminishing returns. Increasing $K$ by a factor of 5 from \num{1000} to \num{5000} only improves Mandarin tone representation by a modest amount (F1 score, defined in Section \ref{evaluation} from 0.80 to 0.87; Figure~\ref{fig:kmeans}). $K$ corresponds to the vocabulary size employed by the downstream language model: increasing $K$ to a high value is not a viable solution.


In this work, we explore three possible routes to a solution: 1) frame-level neural vector quantisers, trained with a \textit{reconstruction} objective, because this might preserve more tone-related variation than K-means; 2) segment-level approaches that take temporal context into account and so might capture tone trajectories; 3) hierarchical or residual `coarse-to-fine' approaches, which might capture phonetic and tonal variation within separate levels.

\section{Method}
\label{sec:methodology}

Our method involves extracting SSL representations from pre-trained foundation models, quantising them using various methods, then probing for both phonetic and tonal information.



\subsection{Data}

We selected data from two representative tone languages: Mandarin (AISHELL-1, 170 h, 400 speakers \cite{bu2017aishell}), and Yorùbá (BibleTTS, 93 h, single speaker \cite{meyer2022bibletts}). All audio was resampled to \SI{16}{\kilo\hertz} to suit the SSL models.
Lexical tone in both languages is realised primarily on the vowel nucleus, though for different phonological reasons. In Mandarin, tones are associated with the syllable as a whole but are phonetically realised primarily on the vowel nucleus \cite{yip2002tone, duanmu2007phonology}. Yorùbá, by contrast, is a canonical vowel-tone language in which the vowel functions as the tone-bearing unit \cite{adeniyi2020lexicalisation,laniran2003downstep}. 

Given this shared vowel-centred realisation of tone, we extract vowel-aligned segments from phonetic alignments for both languages. We follow the official training, validation, and test splits of each corpus when training our probing classifiers.

\subsection{SSL models}

We extract continuous frame-level representations using two HuBERT-based self-supervised models chosen to match the linguistic properties of our target languages. For Mandarin, we use MandarinHuBERT,\footnote{\url{https://huggingface.co/TencentGameMate/chinese-hubert-base}} a HuBERT-based model trained on large-scale Mandarin speech. For Yorùbá, we use AfriHuBERT \cite{alabi25_interspeech}, an extension of mHuBERT-147 trained on speech from 1,226 African languages, including several tone languages such as Yorùbá. This ensures that the representations capture both phonetic detail and tone-relevant information prior to quantisation.
Some of our experimental conditions require phone-level segmentation. For this, we obtain phonetic alignments using the Montreal Forced Aligner (MFA) \cite{mcauliffe17_interspeech} together with language-specific lexica generated using Yoruba-G2P, a tone-aware grapheme-to-phoneme converter for Yorùbá \cite{osakuade2025yorubag2p}. These alignments are used for segmentation and evaluation.





\subsection{Frame-based vs. Segment-based Representations}

Each vowel phone in the dataset is initially represented by a sequence of latent vectors extracted from the SSL model, with one vector per acoustic frame.  We optionally apply mean pooling over the frames aligned to each vowel segment (based on phonetic alignment), resulting in a single vector per segment.

\subsection{Probing classifiers}
\label{evaluation}

Probing is the established task-agnostic method for analysing SSL representations  \cite{belinkov2019analysis, hewitt2019structural}, consistent with benchmarks such as SUPERB \cite{yang21c_interspeech} and SpeechGLUE \cite{ashihara23_interspeech}. The probe is a proxy intended to be a predictor of eventual downstream task performance. Our probes are intentionally simple and lightweight classifiers, and they probe for phone identity and tone only on the vowels (as found using forced alignment). Note that the Mandarin tone probe is for \textit{lexical} tone, because AISHELL-1 does not annotate tone sandhi. 
The type of probe depends on whether each vowel is represented as a sequence (frame-based) or a single vector (segmental). The frame-based probe is an LSTM (1-layer, hidden size 128), followed by task-specific linear heads for phone and tone prediction. The model is trained with $d = 0.3$ dropout with class-weighted cross-entropy loss and early stopping based on validation F1. For segmental features, we train a logistic regression classifier with a softmax output layer. We use cross-entropy loss and evaluate with weighted F1 score. 


The result of quantisation is a sequence of codes (integer indexes into a codebook). Each code has an underlying vector in latent space (e.g., the K-means cluster centroid). We probe these vectors, not the integer codes.


\section{Quantisation Methods}
\label{sec:methods}


The quantisation method is the only experimental variable. Data are fixed and SSL models are frozen. Any variations in the information found by the probes can be attributed to the quantisation strategy alone.

\subsection{Classic K‑means (Frame‑level clustering)}
Our baseline applies standard K-means clustering directly to frame-level latent vectors, which is the method used in the original paper \cite{hsu2021hubert} and is very widely used in the literature. K-means results in a set of $K$ centroids (i.e., vectors in latent space). Each frame in the dataset is then replaced with the closest centroid based on Euclidean distance. K-means is very simple and only uses Euclidean distance in latent space to find clusters.

 
\subsection{Neural Vector Quantisation}

Neural approaches to Vector Quantisation (VQ) use a reconstruction objective, which contrasts with K-means simple distance-based clustering.
We trained neural quantisers following the RepCodec architecture \cite{huang2024repcodec}, where a small encoder–quantiser–decoder is trained to reconstruct the latents. One model performs VQ with a single 500-entry codebook; the other uses Residual VQ (RVQ), with two variants (250 codes $\times$ 2 levels; or $125\times4$). After training, we retain the encoder and quantiser components to discretise SSL representations into DSUs.





\subsection{SVC (Segmentation‑Variant Codebooks).}
Inspired by prior research \cite{sanders25_interspeech}, we assign each frame two centroids: one from a frame-level codebook and one from the corresponding phone-level (pooled) codebook. K-means clustering is used for both. These centroids are then averaged to produce a fused representation that incorporates both frame-level fine-grained, and segmental context.

\subsection{Residual K‑means}
Here we attempt to factor out phonetic information before representing tonal information. First, we run K‑means on mean‑pooled phone segments with coarse quantisation of $K=50$, chosen to roughly match the number of phonemes in Mandarin and Yorùbá. This produces a `phone‑identity' codebook that encodes (only) segmental information. Next, for every original latent vector, we subtract its phone‑level centroid. The remaining `residual' latent is intended to contain less phonetic information than the original latents. We then cluster these residuals with $K=450$. The total number of codes is $50+450=500$, so it is comparable to other experimental conditions.







\begin{table*}[t!]
\caption{F1 scores for phone and tone classification at $K=500$ using different quantisers for Mandarin and Yorùbá. Multi-codebook quantisers (RVQ, SVC, and Residual KMeans) improve tone retention over single-codebook FSQ. Bolded scores are top-2 for tone in each language.}
\centering
\begin{adjustbox}{max width=0.8\linewidth}
\begin{tabular}{l|c|c c|cc}
\hline
\textbf{Quantizer} & \textbf{Levels} & \textbf{Mandarin} &  & \textbf{Yorùbá}\\
& & Phone & Tone & Phone & Tone \\
\hline
\textit{Latent (continuous)}& -- & 0.99 & \textbf{0.94} & 0.97 & \textbf{0.92} \\
\hline
Classic K-means (500)& 1 & 0.99 & 0.70 & 0.95 & 0.77 \\
VQ (500)& 1 & 0.97 & 0.78 &  0.85 & 0.66 \\
RVQ (250x2)& 2 & 0.98 & 0.81 & 0.92 & 0.74 \\
RVQ (125x4)& 4 &  0.99 & \textbf{0.82} & 0.94 & 0.76 \\
\hline

Mean-pooled K-means & 1 & 0.97 & 0.57 & 0.93 & 0.76 \\
SVC (250x2) & 2 & 0.99  & 0.62  & 0.93&0.78 \\
Residual K-means(frame-level) (50+450) & 2 & 0.98 & 0.76 & 0.92& 0.80 \\
Residual K-means(Segmental) (50+450) & 2 & 0.99 & 0.79 & 0.94 & \textbf{0.83}\\
\hline
\end{tabular}
\end{adjustbox}
\label{tab:f1_quantizers_multilingual}
\end{table*}

\begin{figure}[t]
    \centering
    \includegraphics[width=0.5\textwidth]{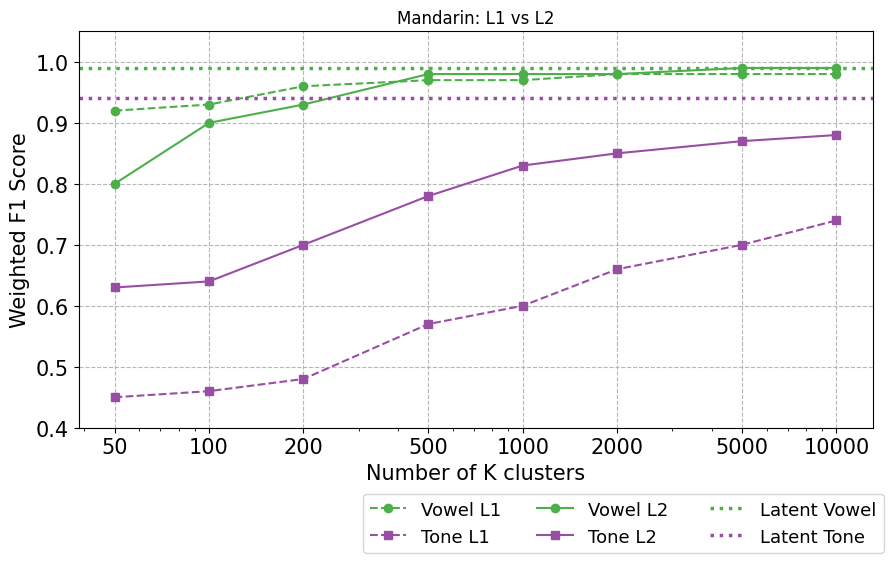} 
    \caption{L1 refers to first-pass K-means on the mean-pooled phone segment at different K; L2 represents clustering on residuals. While L1 performs well for vowels, L2 boosts tone classification, getting closer to the original (unquantised) latents. Dotted lines indicate latent baselines.}
    \label{fig:residual}
\end{figure}




\section{Results and Discussion}

We use probing to evaluate how well each quantisation method preserves phonetic and tonal information.

 
\subsection{Classic K-means degrades tone information}

We find consistently that \textbf{quantisation tends to degrade tone more than phone.} While SSL latents yield near-ceiling F1 scores for both phone and tone classification (e.g., 0.99 / 0.94 on Mandarin), classic frame-level K-means reduces tone F1 to 0.70 (i.e., by 24\%) while phone F1 remains high (Table \ref{tab:f1_quantizers_multilingual}). This reinforces our belief that phonetic information in SSL latent space has much higher variance than tonal information and so dominates during quantisation, whether that is distance-based (K-means) or reconstruction-based (e.g., neural VQ).

Increasing codebook size helps, but only very gradually. As we can see in Figure~\ref{fig:kmeans}, tone F1 score improves very slowly as $K$ increases, but still appears to plateau at an F1 score that is below that for the original latents, even with \num{10000} clusters.

\subsection{Neural VQ is comparable to K-means}

Neural VQ provides improvement over K-means for Mandarin (0.78 for VQ compared to 0.70 for K-means) but not Yorùbá (0.66 for VQ compared to 0.77 for K-means) -- Table~\ref{tab:f1_quantizers_multilingual}. 

\subsection{Neural RVQ retains tone better than K-means}

RVQ performs substantially better than VQ. The deeper configuration (125$\times$4) yields the best retention of tonal information (0.82 for Mandarin, 0.76 for Yorùbá), suggesting that multi-level residual modelling helps preserve tone-related variation in the latent space that a single flat codebook struggles to capture (even with very large codebooks: Figure~\ref{fig:kmeans}). The shallower variant (250$\times$2) shows a similar trend but with a slightly lower tone F1 score.

\subsection{Segment-level clustering}


Mean-pooled phone clustering performs noticeably worse (0.57 for Mandarin and 0.76 for Yorùbá) compared to classic K-means, showing that mean pooling reduces access to pitch trajectories that unfold within phone segments. The SVC method sits between these two extremes but still underperforms the neural methods and, in particular, the residual approach discussed next.

\subsection{Residual K-means}

Our residual K-means formulation aims to factorise phonetic and tonal information by first clustering mean-pooled latents and subsequently clustering the residual variation. This separation yields the best tone retention among the non-neural approaches and is competitive with the best neural method (RVQ). Tone F1 reached 0.79 for Mandarin (a little below the 0.82 for RVQ) and 0.83 for Yorùbá (above the 0.76 for RVQ).

Figure~\ref{fig:residual} provides further insight. The first-pass K-means (`level one', denoted L1) applied to mean-pooled + quantised latents encodes phone well, but not tone. The second-pass clustering on residuals (L2) results in a representation that captures tone much better, getting back towards the upper-bound F1 that we get with the original latents. Again, this pattern of results is consistent with tonal variation being much smaller in magnitude than phonetic variation in the latent space. Accounting for some of the phonetic information with one round of clustering reduces its dominance in the second round.

\subsection{Cross-Language Comparison}

Despite typological differences between Mandarin and Yorùbá, our results show similar trends for both languages: latent features encode both phone and tone very well indeed, but all forms of discretisation degrade the representation of tone. This degradation is most severe for flat, single-codebook methods such as classic K-means and VQ. Quantisation also impacts the two languages differently. Mandarin tone shows a larger drop relative to the continuous latent baseline (0.94 to 0.70 with Classic K-means) than Yorùbá (0.92 to 0.77), suggesting that Mandarin tones are more sensitive to discretisation.

Multi-level quantisers consistently alleviate this degradation, although the most effective approach differs across languages. For Mandarin, deeper RVQ substantially improves tone retention, with RVQ (125×4) achieving the highest quantised tone score (0.82), representing the strongest gain over single-level baselines. For Yorùbá, however, Residual K-means (Segmental) achieves the best tone performance (0.83), surpassing all RVQ variants and demonstrating the value of segment-aware residual modelling. These patterns suggest that Mandarin tone benefits most from multi-level hierarchical quantisation, whereas Yorùbá tone benefits most from segmental residual modelling.

We attribute this to the nature of tone realisation: Yorùbá has stable, vowel-aligned level tones, which align well with phone-based segmentation, making segmental residual modelling especially effective. In contrast, Mandarin’s contour tones, which exhibit greater coarticulation and sandhi effects, may require deeper or more flexible temporal modelling to preserve tonal distinctions.


\section{Limitations}
Our analysis used only representation probing rather than downstream tasks. This is a standard methodology, but ultimately we need to measure downstream task performance.

Our probes used forced alignments. This is not a limitation, since they are only required during evaluation, not quantisation.

However, our Residual K-means approach required those alignments during quantisation. If these were not available (e.g., a low-resource setting) then options would include mean-pooling over fixed-duration segments, or automatic unit discovery. Alternatively, neural RVQ can be used, which does not require alignments.


\section{Conclusion}

This study examined how a range of quantisation strategies represent lexical tone in two typologically distinct tone languages, Mandarin and Yorùbá. While tone is well encoded in the continuous SSL latents, our probing results show that discretisation always degrades tonal information more than segmental information. We believe that this result also has implications for the representation of prosody in SSL latents, and that discretisation will very likely degrade that too.


The degradation can be mitigated by multi-level quantisation: here, we compared only neural RVQ and two rounds of K-means. More sophisticated schemes could surely be devised, which may retain more of the tone information that exists in the original latents.


More broadly, our findings point to an emerging challenge for the design of Discrete Speech Units (DSUs): quantisers implicitly emphasise segmental information and therefore may encode suprasegmental information less well. Tone is one example, but similar issues are likely to arise for prosodic dimensions such as phrasing, prominence, and rhythm. Future work should develop tone-aware or prosody-aware discretisation schemes.


Such developments have implications for speech LLMs and TTS systems, especially when these technologies are applied to under-resourced tone languages, where accurate suprasegmental representation is especially valuable. 
Stronger tone retention is particularly relevant for TTS and speech-to-speech translation in tone languages, where tone errors are lexically contrastive; improvements in tone preservation will directly reduce homophone confusions and improve perceived naturalness.

\section{Acknowledgements}
This  work  was  supported  in part  by  the UKRI Centre  for Doctoral   Training   in Natural Language  Processing,  funded by the UKRI (grant EP/S022481/1) and the University of Edinburgh.

We thank Korin Richmond for his constructive suggestions and detailed review of an earlier draft, which greatly improved the presentation of this work.

\bibliographystyle{IEEEtran}

\bibliography{mybib}


\end{document}